# License Plate Recognition System Based on Color Coding Of License Plates

Jani Biju Babjan
S5 Information Technology, Government Engineering College, Barton Hill, Thiruvananthapuram
janibiju@yahoo.co.in

Abstract
　　License Plate Recognition Systems are used to determine the license plate number of a vehicle. The current system mainly uses Optical Character Recognition to recognize the number plate. There are several problems to this system. Some of them include interchanging of several letters or numbers (letter 'O' with digit '0'), difficulty in localizing the license plate, high error rate, use of different fonts in license plates etc. So a new system to recognize the license plate number using color coding of license plates is proposed in this paper. Easier localization of license plate can be done by searching for the start or stop patters of license plates. An eight segment display system along with traditional numbering with the first and last segments left for start or stop patterns is proposed in this paper. Practical applications include several areas under Internet of Things (IoT).

Keywords: License plate recognition, Color coding, Internet of Things

Introduction
　　The Automatic Number Plate Recognition (ANPR) system which is currently used makes use of Optical Character Recognition. There are several algorithms that are needed for the accurate working of this system. First the plate needs to be localized in the image. Then if the plate is angled, the skew needs to be adjusted. The image is normalized and adjusted for brightness and contrast. The characters are then segmented and optical character recognition is applied to get the number. But several problems are present. Any mistake in one of the algorithms can lead to several mistakes in the final result thereby reducing the accuracy of the system. These traditional algorithms rely heavily on the high contrast of the characters that appear on a number plate (black characters on white or yellow background)

　　So a new system, one which uses color coded license plates can be used for recognizing the license plate. The system first converts the license plate number which is in base-36 to base-255. For each number in base-255 scheme, there is a color which corresponds to the number. The color is then used to color the license plate. Refer the tables 1 and 2 for details of the scheme being used for the conversion.

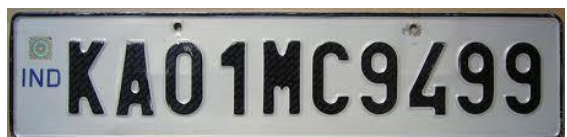
A standard number plate that is used in India

There are eight segments present in the license plate. The first segment is reserved for the start pattern, segments two to seven hold the color code of the license plate number and the eighth segment holds the end pattern. The start and end patterns contains three bands of red, blue and green colors (with 255 as the value in the respective rgb column) respectively.

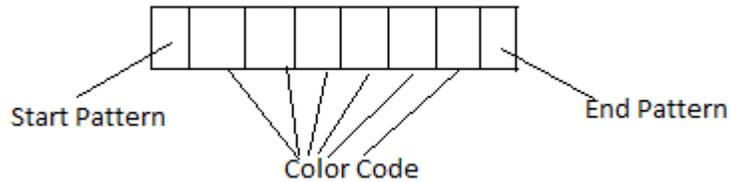

The major algorithms that are used in the system include template matching which localizes the plate, adjustment for angled plates and conversion between the number systems that are used to identify the plates. By using the color coding, the license plate numbers can be easily identified.

**Methods Used**

The system assumes a license plate number as a little endian base-36 number. Characters like space and hyphen are removed from the number. The two sections below discuss ways to generate the color code taking a license plate as input and recognizing the license plate number by taking an image containing the color code as input.

**1. Generation of Color code**

In order to generate the color code to be used in the vehicle, we give the license plate number of the vehicle as the input. The license plate number is considered to be a little endian base-36 number. First, the little endian base-36 number is converted into decimal. After conversion, the decimal number is converted to base-255 format. Then the base-255 format is used to generate the color code, according to the table-2.The procedure to generate the color code is given below.

Procedure:
1. Read the license plate number.
2. Convert the number to decimal considering it as a little endian base-36 number.
3. Now convert the decimal number to base-255 scheme, the value returned is an array.
4. Color the plate using the color corresponding to each of the numbers in the array.
5. Attach the start and stop patterns to the image.
6. Also write the number in the traditional way to the license plate.

The number is also written in the traditional method so that humans can also understand the number, in case it is found in an accident or on a crime scene.

**2. Recognition of Color Code**

The recognition of the color code which is present in an image is done by using template matching algorithms. Template matching is used in the image to find the position of the regions that mostly resemble the start/end patterns. Then, the rgb color composition of pixels in the region bounded by start/end patterns is found. Each segment of the code is identified using some approximation techniques for finding the length of a segment in the image and color value is obtained for all segments. The color values are then converted to their base-255 equivalent using the values in the table-2. Now we need to reverse the process used to generate a color code. The base-255 number is converted to decimal and then the decimal is converted back to base-36 to obtain the license plate number.

# Results Obtained
## 1. Generating the color code

    The following color codes were generated for the corresponding license plate numbers according to a program written in Python using Python Imaging Library (PIL). The number can be printed to the plate using a color that can be distinguished from the color of the blocks in the plate.

| License Plate Number | Generated Color Code |
| --- | --- |
| KL29H5643 | 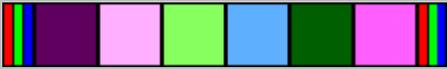 |
| KL04Q9399 | 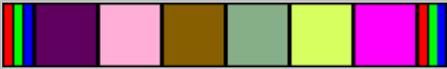 |
| KA01MC949 | 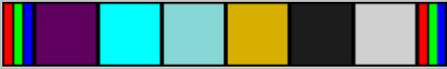 |
| PBX2384 | 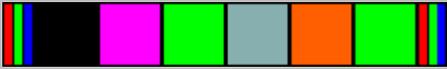 |
| SK01GO693 | 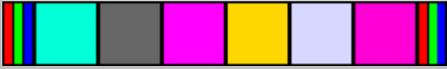 |

    A completed license plate (along with the number) will appear as shown below. The plate is computer generated as the rest of the images. In this image white (hex 0xffffff) is used to print the number on the plate as it can be distinguished from the colors used in blocks.

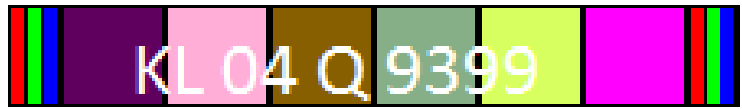

## 2. Recognizing the color code

A simulator was created to obtain these results. The following images show the result after applying the template matching algorithm. These results were obtained by experimenting on computer generated images with Python and OpenCV library on a Windows 8 computer . The template matching algorithm is used to search for the start and stop patterns. Both are matched against the template that is shown below:

Template for matching the start and stop pattern

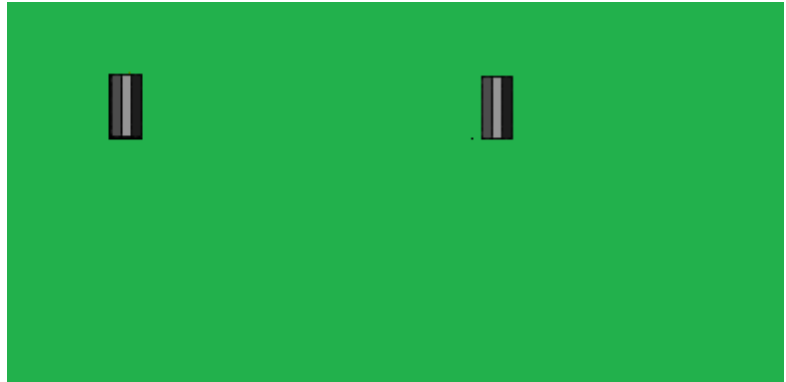

Resultant image after applying Template Matching

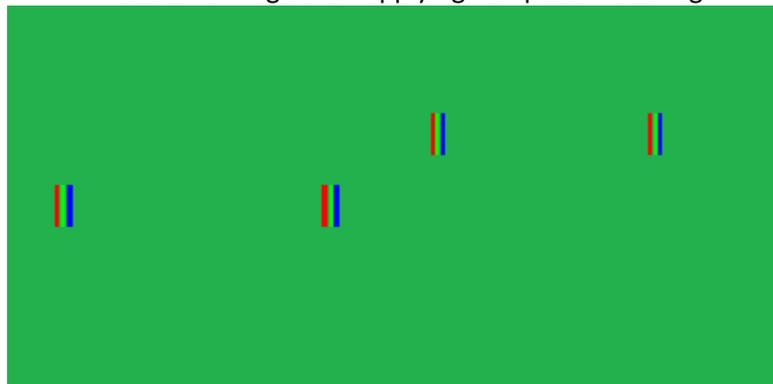

Resultant Image after applying template matching containing multiple codes

The results obtained after matching the template, reading the pixels and converting to the original license plate number is shown in the table below :

| Image(Reduced in size) | Actual Resolution | Actual License Plate Number |
|---|---|---|
| 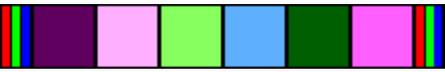 (Originally Generated Image) | 224 X 32 px | KL29H5643 |
| 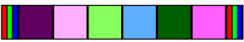 (On a white background) | 400 X 120 px | KL29H5643 |
| 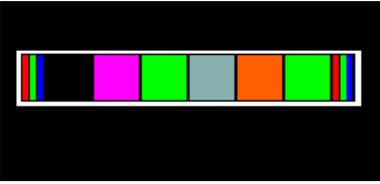 (On a black background) | 256 X 120 px | SK01GO693 |
| 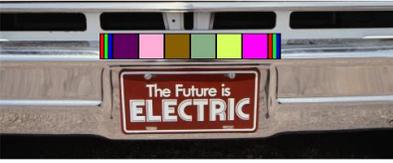 (On a random image) | 500 X 200 px | KL04Q9399 |
| 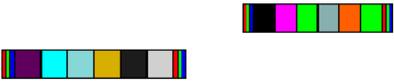 | 400 X 200 px | PBX2384 KA01MC949 |

**Observations**

The following observations were obtained after experimenting with the computer generated images using the simulator:
1. A high accuracy rate was observed in the case of computer generated images.
2. Using the methods specified in the simulator few of the errors that occurred were corrected
3. Pattern recognition can be used to identify even multiple codes in a single image and accurate results were obtained.

## Advantages

The system has a few advantages over the traditional Automatic Number Plate Recognition (ANPR) systems that use Optical Character Recognition (OCR). They are listed below:

1. The system uses algorithms that take smaller time compared to the traditional ANPR systems that use OCR.
2. The current system has to process the image, convert it to grey-scale, adjust for contrast or brightness, identify the number plate in the picture using some segmentation techniques, extract the license plate, divide the characters to segments and finally apply OCR to get the result. Here only a pattern recognition (for start and end) and color look-up is needed to find the result.
3. No changes have to be made in the image.
4. Large number of number plates can be accommodated by the system, even the six segment code can be used to represent approximately $10^{14}$ numbers

## Applications

The major application of the system comes under Internet of Things (IoT). Few of the practical applications of the system include:

1. Traffic Control and violations: The system can be used to identify vehicles that exceed a given speed limit, create an accident, identify stolen vehicles and detect vehicles parked in the wrong side.
2. Collection of toll or parking fees: The recognition system can be used to identify the vehicle and find the toll/parking fee which needs to be collected.
3. Targeted advertising: We can identify the vehicle, find the owner and give his preferences as the advertisement in the bill board.
4. Universal Database of Number plates: A universal database can be created to hold and easily look up the numbers
5. The color code can be extended to other objects as well as a code to identify the things. More data can be stored in the color code with the advent of better technologies.

## Challenges

There are also some problems that the system needs to tackle. Some of them are listed below:

1. Tampering in the license plate: If the license plate was covered in say, dirt or some other material, the system fails to identify the number.
2. The current accuracy and cost of printing and photographic technologies may prevent the implementation of the system in a large scale.

## Discussions And Future Work

This paper has presented a simple method for license plate recognition. Real time applications require a highly accurate and high speed camera, generation (or printing) of exact color to be used in the license plate etc. Real time application also requires a very quick processing of the image. There is also plan to include a parity check using one more segment. The system can also be extended to be a universal system that can recognize other products as well. This requires more expansion of the algorithms to generate the code.

**References**
1. S.Kranthi, K.Pranathi, A.Srisaila "Automatic Number Plate Recognition", 2011
2. http://en.wikipedia.org/ "Template Matching"
3. http://en.wikipedia.org/ "Automatic Number Plate Recognition"

**Conversion Tables**
Base-36 to Decimal Values(Table -1)

| Character | Value in base-36 | Character | Value in base-36 |
|---|---|---|---|
| 0 | 0 | I/i | 18 |
| 1 | 1 | J/j | 19 |
| 2 | 2 | K/k | 20 |
| 3 | 3 | L/l | 21 |
| 4 | 4 | M/m | 22 |
| 5 | 5 | N/n | 23 |
| 6 | 6 | O/o | 24 |
| 7 | 7 | P/p | 25 |
| 8 | 8 | Q/q | 26 |
| 9 | 9 | R/r | 27 |
| A/a | 10 | S/s | 28 |
| B/b | 11 | T/t | 29 |
| C/c | 12 | U/u | 30 |
| D/d | 13 | V/v | 31 |
| E/e | 14 | W/w | 32 |
| F/f | 15 | X/x | 33 |
| G/g | 16 | Y/y | 34 |
| H/h | 17 | Z/z | 35 |

Color Code Conversion Values (Table-2)

| Value | Color Code(Hex) | Value | Color Code(Hex) | Value | Color Code(Hex) |
|---|---|---|---|---|---|
| 0 | 0x000000 | 85 | 0x5fffaf | 170 | 0xd75fd7 |
| 1 | 0x800000 | 86 | 0x5fffd7 | 171 | 0xd75fff |
| 2 | 0x008000 | 87 | 0x5fffff | 172 | 0xd78700 |
| 3 | 0x808000 | 88 | 0x870000 | 173 | 0xd7875f |
| 4 | 0x000080 | 89 | 0x87005f | 174 | 0xd78787 |
| 5 | 0x800080 | 90 | 0x870087 | 175 | 0xd787af |
| 6 | 0x008080 | 91 | 0x8700af | 176 | 0xd787d7 |
| 7 | 0xc0c0c0 | 92 | 0x8700d7 | 177 | 0xd787ff |
| 8 | 0x808080 | 93 | 0x8700ff | 178 | 0xd7af00 |
| 9 | 0xff0000 | 94 | 0x875f00 | 179 | 0xd7af5f |
| 10 | 0x00ff00 | 95 | 0x875f5f | 180 | 0xd7af87 |
| 11 | 0xffff00 | 96 | 0x875f87 | 181 | 0xd7afaf |
| 12 | 0x0000ff | 97 | 0x875faf | 182 | 0xd7afd7 |
| 13 | 0xff00ff | 98 | 0x875fd7 | 183 | 0xd7afff |
| 14 | 0x00ffff | 99 | 0x875fff | 184 | 0xd7d700 |
| 15 | 0xffffff | 100 | 0x878700 | 185 | 0xd7d75f |
| 16 | 0x0f0f0f | 101 | 0x87875f | 186 | 0xd7d787 |
| 17 | 0x00005f | 102 | 0x878787 | 187 | 0xd7d7af |
| 18 | 0x000087 | 103 | 0x8787af | 188 | 0xd7d7d7 |
| 19 | 0x0000af | 104 | 0x8787d7 | 189 | 0xd7d7ff |
| 20 | 0x0000d7 | 105 | 0x8787ff | 190 | 0xd7ff00 |
| 21 | 0x0000ff | 106 | 0x87af00 | 191 | 0xd7ff5f |
| 22 | 0x005f00 | 107 | 0x87af5f | 192 | 0xd7ff87 |
| 23 | 0x005f5f | 108 | 0x87af87 | 193 | 0xd7ffaf |
| 24 | 0x005f87 | 109 | 0x87afaf | 194 | 0xd7ffd7 |
| 25 | 0x005faf | 110 | 0x87afd7 | 195 | 0xd7ffff |
| 26 | 0x005fd7 | 111 | 0x87afff | 196 | 0xff0000 |
| 27 | 0x005fff | 112 | 0x87d700 | 197 | 0xff005f |
| 28 | 0x008700 | 113 | 0x87d75f | 198 | 0xff0087 |
| 29 | 0x00875f | 114 | 0x87d787 | 199 | 0xff00af |
| 30 | 0x008787 | 115 | 0x87d7af | 200 | 0xff00d7 |
| 31 | 0x0087af | 116 | 0x87d7d7 | 201 | 0xff00ff |
| 32 | 0x0087d7 | 117 | 0x87d7ff | 202 | 0xff5f00 |
| 33 | 0x0087ff | 118 | 0x87ff00 | 203 | 0xff5f5f |
| 34 | 0x00af00 | 119 | 0x87ff5f | 204 | 0xff5f87 |
| 35 | 0x00af5f | 120 | 0x87ff87 | 205 | 0xff5faf |
| 36 | 0x00af87 | 121 | 0x87ffaf | 206 | 0xff5fd7 |
| 37 | 0x00afaf | 122 | 0x87ffd7 | 207 | 0xff5fff |
| 38 | 0x00afd7 | 123 | 0x87ffff | 208 | 0xff8700 |
| 39 | 0x00afff | 124 | 0xaf0000 | 209 | 0xff875f |
| 40 | 0x00d700 | 125 | 0xaf005f | 210 | 0xff8787 |
| 41 | 0x00d75f | 126 | 0xaf0087 | 211 | 0xff87af |
| 42 | 0x00d787 | 127 | 0xaf00af | 212 | 0xff87d7 |
| 43 | 0x00d7af | 128 | 0xaf00d7 | 213 | 0xff87ff |
| 44 | 0x00d7d7 | 129 | 0xaf00ff | 214 | 0xffaf00 |
| 45 | 0x00d7ff | 130 | 0xaf5f00 | 215 | 0xffaf5f |
| 46 | 0x00ff00 | 131 | 0xaf5f5f | 216 | 0xffaf87 |
| 47 | 0x00ff5f | 132 | 0xaf5f87 | 217 | 0xffafaf |
| 48 | 0x00ff87 | 133 | 0xaf5faf | 218 | 0xffafd7 |
| 49 | 0x00ffaf | 134 | 0xaf5fd7 | 219 | 0xffafff |
| 50 | 0x00ffd7 | 135 | 0xaf5fff | 220 | 0xffd700 |
| 51 | 0x00ffff | 136 | 0xaf8700 | 221 | 0xffd75f |
| 52 | 0x5f0000 | 137 | 0xaf875f | 222 | 0xffd787 |
| 53 | 0x5f005f | 138 | 0xaf8787 | 223 | 0xffd7af |
| 54 | 0x5f0087 | 139 | 0xaf87af | 224 | 0xffd7d7 |
| 55 | 0x5f00af | 140 | 0xaf87d7 | 225 | 0xffd7ff |
| 56 | 0x5f00d7 | 141 | 0xaf87ff | 226 | 0xffff00 |
| 57 | 0x5f00ff | 142 | 0xafaf00 | 227 | 0xffff5f |
| 58 | 0x5f5f00 | 143 | 0xafaf5f | 228 | 0xffff87 |
| 59 | 0x5f5f5f | 144 | 0xafaf87 | 229 | 0xffffaf |
| 60 | 0x5f5f87 | 145 | 0xafafaf | 230 | 0xffffd7 |
| 61 | 0x5f5faf | 146 | 0xafafd7 | 231 | 0xffffff |
| 62 | 0x5f5fd7 | 147 | 0xafafff | 232 | 0x080808 |
| 63 | 0x5f5fff | 148 | 0xafd700 | 233 | 0x121212 |
| 64 | 0x5f8700 | 149 | 0xafd75f | 234 | 0x1c1c1c |
| 65 | 0x5f875f | 150 | 0xafd787 | 235 | 0x262626 |
| 66 | 0x5f8787 | 151 | 0xafd7af | 236 | 0x303030 |
| 67 | 0x5f87af | 152 | 0xafd7d7 | 237 | 0x3a3a3a |
| 68 | 0x5f87d7 | 153 | 0xafd7ff | 238 | 0x444444 |
| 69 | 0x5f87ff | 154 | 0xafff00 | 239 | 0x4e4e4e |
| 70 | 0x5faf00 | 155 | 0xafff5f | 240 | 0x585858 |
| 71 | 0x5faf5f | 156 | 0xafff87 | 241 | 0x606060 |
| 72 | 0x5faf87 | 157 | 0xafffaf | 242 | 0x666666 |
| 73 | 0x5fafaf | 158 | 0xafffd7 | 243 | 0x767676 |
| 74 | 0x5fafd7 | 159 | 0xafffff | 244 | 0x808080 |
| 75 | 0x5fafff | 160 | 0xd70000 | 245 | 0x8a8a8a |
| 76 | 0x5fd700 | 161 | 0xd7005f | 246 | 0x949494 |
| 77 | 0x5fd75f | 162 | 0xd70087 | 247 | 0x9e9e9e |

| 78 | 0x5fd787 | 163 | 0xd700af | 248 | 0xa8a8a8 |
| 79 | 0x5fd7af | 164 | 0xd700d7 | 249 | 0xb2b2b2 |
| 80 | 0x5fd7d7 | 165 | 0xd700ff | 250 | 0xbcbcbc |
| 81 | 0x5fd7ff | 166 | 0xd75f00 | 251 | 0xc6c6c6 |
| 82 | 0x5fff00 | 167 | 0xd75f5f | 252 | 0xd0d0d0 |
| 83 | 0x5fff5f | 168 | 0xd75f87 | 253 | 0xdadada |
| 84 | 0x5fff87 | 169 | 0xd75faf | 254 | 0xe4e4e4 |